\documentclass[journal]{IEEEtran}
\usepackage{amsmath,amsfonts}
\usepackage{algorithmic}
\usepackage{algorithm}
\usepackage{array}
\usepackage[caption=false,font=normalsize,labelfont=sf,textfont=sf]{subfig}
\usepackage{textcomp}
\usepackage{stfloats}
\usepackage{url}
\usepackage{verbatim}
\usepackage{graphicx}
\usepackage{makecell}
\usepackage{cite}
\graphicspath{{./img/}}
\begin{document}

\title{Knowledge-Base based Semantic Image Transmission Using CLIP}

\author{Chongyang Li, Yanmei He, Tianqian Zhang, Mingjian He, and Shouyin Liu
\thanks{Chongyang Li, Yanmei He, Tianqian Zhang, Mingjian He and
	Shouyin liu are with the Dept. of Electronic \& information Engineering, College of Physical Science and Technology, Central China Normal University, Wuhan 430079, China (email: lichongyang2020@mails.ccnu.edu.cn; 261ym\_He@mails.ccnu.edu.cn;
	tqzhang@mails.ccnu.edu.cn; mingjian.he@ccnu.edu.cn; syliu@ccnu.edu.cn)\\
	This work was supported in part by  the Key Research and
	Development Program of Hubei Province under Grant 2023BAB061,
	and in part by China Postdoctoral Science Foundation under Grant
	2023M741315, and in part by Postdoctor Project of Hubei Province under
	Grant 2024HBBHCXA049.\\ Code are available at https://github.com/Linkcy97/KBSC }
}



\maketitle

\begin{abstract}
This paper proposes a novel knowledge-Base (KB) assisted semantic communication framework for image transmission. At the receiver, a Facebook AI Similarity Search (FAISS) based vector database is constructed by extracting semantic embeddings from images using the Contrastive Language-Image Pre-Training (CLIP) model. During transmission, the transmitter first extracts a 512-dimensional semantic feature using the CLIP model, then compresses it with a lightweight neural network for transmission. After receiving the signal, the receiver reconstructs the feature back to 512 dimensions and performs similarity matching from the KB to retrieve the most semantically similar image. Semantic transmission success is determined by category consistency between the transmitted and retrieved images, rather than traditional metrics like Peak Signal-to-Noise Ratio (PSNR). The proposed system prioritizes semantic accuracy, offering a new evaluation paradigm for semantic-aware communication systems. Experimental validation on CIFAR100 demonstrates the effectiveness of the framework in achieving semantic image transmission.
\end{abstract}


\section{Introduction}
\IEEEPARstart{R}{e}cent advances in deep learning and semantic communication have shifted the focus of communication from traditional bit-level data transmission to semantic-aware delivery, where the goal is to preserve the meaning of transmitted content rather than its exact reconstruction\cite{luo2022semantic}. In the field of image transmission, the efficient transmission of images in bandwidth constrained and noisy environments is a fundamental challenge for modern communication systems. Traditional image transmission methods rely on source-channel coding (SCC) techniques, such as JPEG compression followed by channel coding, to ensure image quality at the receiver. Traditional image transmission methods rely on source-channel coding techniques, such as JPEG source coding to remove redundant compressed data, followed by channel coding to ensure bit-level accuracy. SCC evaluation metrics still focus on bit error rate (BER), where the successful decoding of an image depends on whether the erroneous bits occur in critical parts of the JPEG data.

\begin{figure*}[ht]
	\centering
	\includegraphics[width=7.5in]{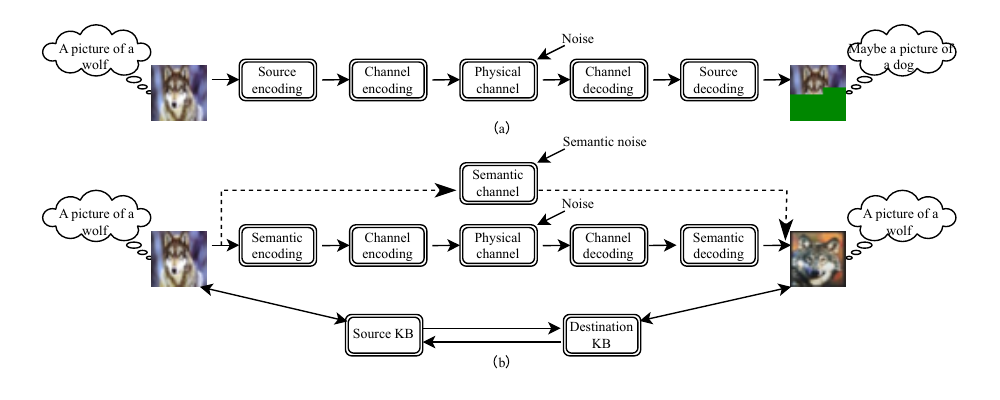}
	\caption{The comparison between traditional image transmission and semantic image transmission}
	\label{fig_1}
\end{figure*}

As shown in Fig.\ref{fig_1}(a), traditional methods rely on channel encoding to mitigate the impact of noise. However, when errors occur in critical regions of the source encoded data, the reconstructed image may be severely degraded, leading to misinterpretation of its content. For example, in this scenario, the original image of a wolf is received as a distorted version, which is mistakenly perceived as a dog. In contrast, semantic image transmission aims to preserve high-level meaning rather than pixel-level accuracy. As illustrated in Fig\ref{fig_1}(b), the transmitter extracts and transmits semantic features rather than raw pixel values. Original image undergo semantic encoding, channel encoding, and transmission through a physical channel before being decoded and matched with the Knowledge-Base (KB) at the receiver. This process introduces semantic noise, which different from conventional channel noise. Semantic noise can arise from two main sources: 1) KB mismatch: Even in the absence of channel noise, if the transmitter and receiver possess different KBs, the receiver may fail to correctly interpret the transmitted semantics. 2) Excessive channel noise: When the transmitter and receiver share a same KB, severe channel noise can still distort the transmitted semantic features to the extent that they no longer accurately represent the original content. This leads to semantic misinterpretation.

With the widespread use of deep learning (DL), bourtsoulatze\cite{bourtsoulatze2019deep} has applied it to joint source-channel coding (JSCC) for image transmission, aiming to enhance transmission performance under low signal-to-noise ratio (SNR) conditions. This JSCC approach directly maps image pixel values to complex-valued channel input signals, enabling image reconstruction even in the presence of errors. However, achieving optimal performance at different SNR levels requires training separate neural networks for each condition. To address this issue, Xu\cite{xu2023deep} designed attention feature module that receives Channel State Information (CSI) to enable adaptation to different SNR levels. Furthermore, \cite{yang2022deep,zhang2023predictive,he2023rate} employ neural networks to achieve variable-rate transmission under different SNR conditions. Their goal is to transmit fewer features in good channel conditions to improve transmission speed while transmitting more features in poor channel conditions to enhance image quality. While above-mentioned methods achieve image transmission under low SNR conditions, they primarily focus on pix-level reconstruction. Consequently, their performance is evaluated using traditional metrics such as Peak Signal-to-Noise Ratio (PSNR) or Multi-Scale Structural Similarity (MS-SSIM)\cite{wang2003multiscale}. However, these metrics fail to align with human perceptual quality or semantic fidelity, as they overly emphasize pixel-wise or low-level structural accuracy. To address these limitations, more advanced perceptual metrics such as Learned Perceptual Image Patch Similarity (LPIPS)\cite{zhang2018unreasonable}, Deep Image Structure and Texture Similarity (DISTS) \cite{ding2020image}, and Fréchet Inception Distance (FID) \cite{heusel2017gans} have been proposed. These metrics better capture perceptual degradations and offer stronger correlations with human judgment.

In order to realize the transmission at the semantic level of images, this paper proposes a KB based semantic image transmission framework using Contrastive Language-Image Pre-Training (CLIP). In this approach, the transmitted data consists of semantic features extracted using CLIP, rather than pixel-level compression features. Upon receiving the transmitted semantic features, the receiver searches a pre-established semantic vector KB to find the most similar image. In our experiments, we construct the KB using images from CIFAR100\cite{krizhevsky2009learning}, where each category has a representative set of images. Practical implementations can dynamically update the KB based on application scenarios. To further compress the features and enhance noise resilience, we design a lightweight encoder-decoder neural network that compresses the 512-dimensional CLIP semantic features while adapting to different SNR conditions. Since traditional image reconstruction metrics are no longer applicable in this framework, we introduce semantic accuracy as a new evaluation metric to assess system performance. Experiments conducted on the CIFAR100 demonstrate that the proposed approach outperforms traditional BPG+LDPC schemes as well as deep learning-based SwinJSCC\cite{yang2024swinjscc} in terms of semantic transmission accuracy.

\section{KB based Semantic Image Transmission}

\subsection{Overall Architecture}
\begin{figure*}[ht]
	\centering
	\includegraphics[width=7.5in]{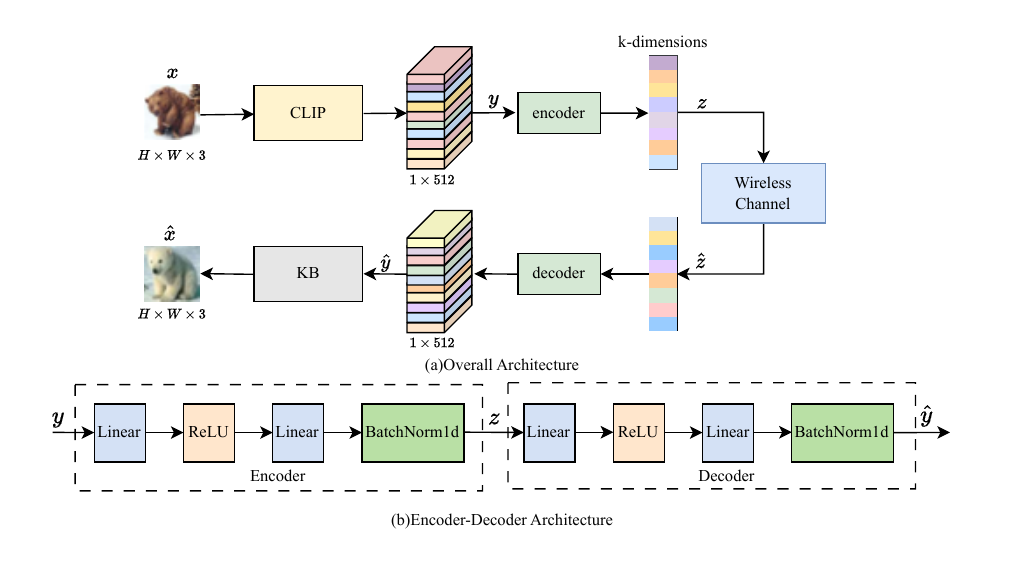}
	\caption{The proposed image semantic transmission framework using CLIP}
	\label{fig_2}
\end{figure*}

The KB based image semantic transmission framework using CLIP is shown in Fig.\ref{fig_2}(a). The whole framework consists of encoder, decoder, CLIP and KB. The KB consists of CLIP vectors $KB = \{kb_1,kb_2...,kb_M\}\in\mathbb{R}^{1\times512}$ at the receiver, $M$ denotes the number of images.  An RGB image $x\in\mathbb{R}^{H\times W\times3}$ is initially fed into CLIP for preliminary extraction of semantic features $y\in\mathbb{R}^{1\times 512}$. Then, the features are compressed by an encoder to get a feature of $z\in\mathbb{R}^{1\times k}$. The compressed features will be modulated into $z\in\mathbb{C}^{1\times \frac{k}{2}}$ signal and transmitted through a wireless channel. The overall transmission cost is measured using the Channel Bandwidth Ratio (CBR), which quantifies the ratio between the number of channel input symbols and the number of source image symbols\cite{yang2024swinjscc}, as shown in $\frac{\frac{k}{2}}{H\times W\times3}$.

The signal transformation can be expressed as $\hat{z}=Hz+n$, where $H$ is the channel gain, and $n$ denotes the additive noise. We consider two types of channel conditions: Gaussian channel and Rayleigh fading channel. In Gaussian channel, the channel gain $H$ remains constant at 1,  while the additive noise $n$ follows a complex Gaussian distribution $n \sim \mathcal{CN}(0,1)$. In Rayleigh fading channel, both the channel gain and the additive noise follow a complex Gaussian distribution $H, n \sim \mathcal{CN}(0,1)$. In our implementation, the attenuation factor $H$ in the Rayleigh channel varies independently for each transmitted symbol, following an independent and identically distributed (i.i.d.) Rayleigh fading model.

After the semantic signal $\hat{z}$ is transmitted to the receiver, a symmetric decoder reconstructs it into a 512-dimensional CLIP semantic feature $\hat{y}$. The reconstructed feature is then compared with the KB using similarity matching to identify the most semantically similar image. Specifically, the matched image $\hat{x}$ is determined by finding the vector $kb_i$ in KB that has the smallest L2 distance to $\hat{y}$, formulated as:

\begin{equation}
	\hat{x_i} = \arg\min_{kb_i \in KB} \|\hat{y} - kb_i\|_2
\end{equation}
The image associated with the closest vector is then retrieved as the final received image $\hat{x_i}$

\subsection{Knowledge Base}
To efficiently retrieve the most semantically similar image at the receiver, we employ a KB consisting of pre-extracted CLIP feature vectors. The KB is constructed by extracting 512-dimensional CLIP embeddings from a dataset of $M$ reference images, denoted as: 
\begin{equation}
	KB = \{kb_1,kb_2...,kb_M\}\in\mathbb{R}^{1\times512}
\end{equation}
where each vector $kb_i$ corresponds to an image $\hat{x_i}$ in the dataset. 

To facilitate fast and scalable similarity search, we utilize Facebook AI Similarity Search (FAISS), an efficient library for nearest neighbor retrieval\cite{johnson2019billion}. FAISS supports both CPU and GPU-based indexing, enabling accelerated search operations that align with the computational capabilities of future wireless devices. This allows for high-speed indexing and querying, which is crucial for real-time image retrieval in wireless communication scenarios. At the receiver, after obtaining the reconstructed CLIP feature$\hat{y}$, the most semantically similar image $\hat{x}$ is determined by performing an L2 distance search in the KB. Using FAISS, this search operation is performed efficiently, even when the KB contains numerous images. The retrieved image is then used as the final output, ensuring that the received image maintains semantic consistency with the transmitted content.

\subsection{Encoder and Decoder}
The semantic features extracted by CLIP inherently exhibit a certain level of robustness to noise, as they capture high-level semantic information rather than raw pixel values. To further enhance noise resilience and reduce transmission overhead, we design a lightweight encoder-decoder network based on a simple Multi-Layer Perceptron (MLP) as shown in Fig.\ref{fig_2}(b).

The encoder consists of two fully connected layers with a ReLU activation in between, followed by a Batch Normalization layer to stabilize feature distribution. It compresses the original 512-dimensional CLIP feature into a $k$-dimensional representation, effectively reducing the transmission cost. The decoder, which mirrors the structure of encoder, reconstructs the received features back to 512 dimensions.

This design achieves a balance between dimensionality reduction and robustness while maintaining low computational complexity. The simplicity of  MLP-based architecture ensures minimal inference latency, making it well-suited for real-time semantic communication applications.

\section{Simulation}
In this section, we trained and evaluated the proposed scheme under different CBR conditions and compared it with other schemes under different SNRs. The CIFAR100 is used for training, validation, and testing. For each CBR in $[\frac{1}{48} , \frac{1}{24} , \frac{1}{12} , \frac{1}{6} , \frac{1}{3}]$, a dedicated model is trained to adapt to different SNRs in $[-7, -4, 0, 4, 7]$ dB and evaluated on a broader range of SNRs in $[-7,-6,-5,-4,-2,0,2,4,5,6,7,10]$. To ensure a fair evaluation, the CIFAR100 dataset is partitioned as follows: The original training set is split into an 8:2 ratio for training and validation. The validation and test sets are further divided and each category is equally split into two subsets, one used as the set of transmitted images at the transmitter and the other as the KB at the receiver.

During training, only the training set is used, while the validation set is employed to select the best perform model. The final evaluation is conducted using the test set. Model performance is measured using semantic accuracy, which is defined as the percentage of transmitted images whose category matches the category of the retrieved image from the KB of receiver. If the matched image belongs to the same category as the transmitted image, the transmission is considered successful; otherwise, it is considered incorrect.

For comparison, the semantic accuracy of other methods is evaluated by first passing the reconstructed images through CLIP to extract semantic features. These extracted features are then matched against the KB using the same similarity retrieval method to determine whether the retrieved image belongs to the correct category. All experiments were performed in WSL with 13600K and 4070TiSuper.

\begin{figure}[h]
	\centering
	\includegraphics[width=3.5in]{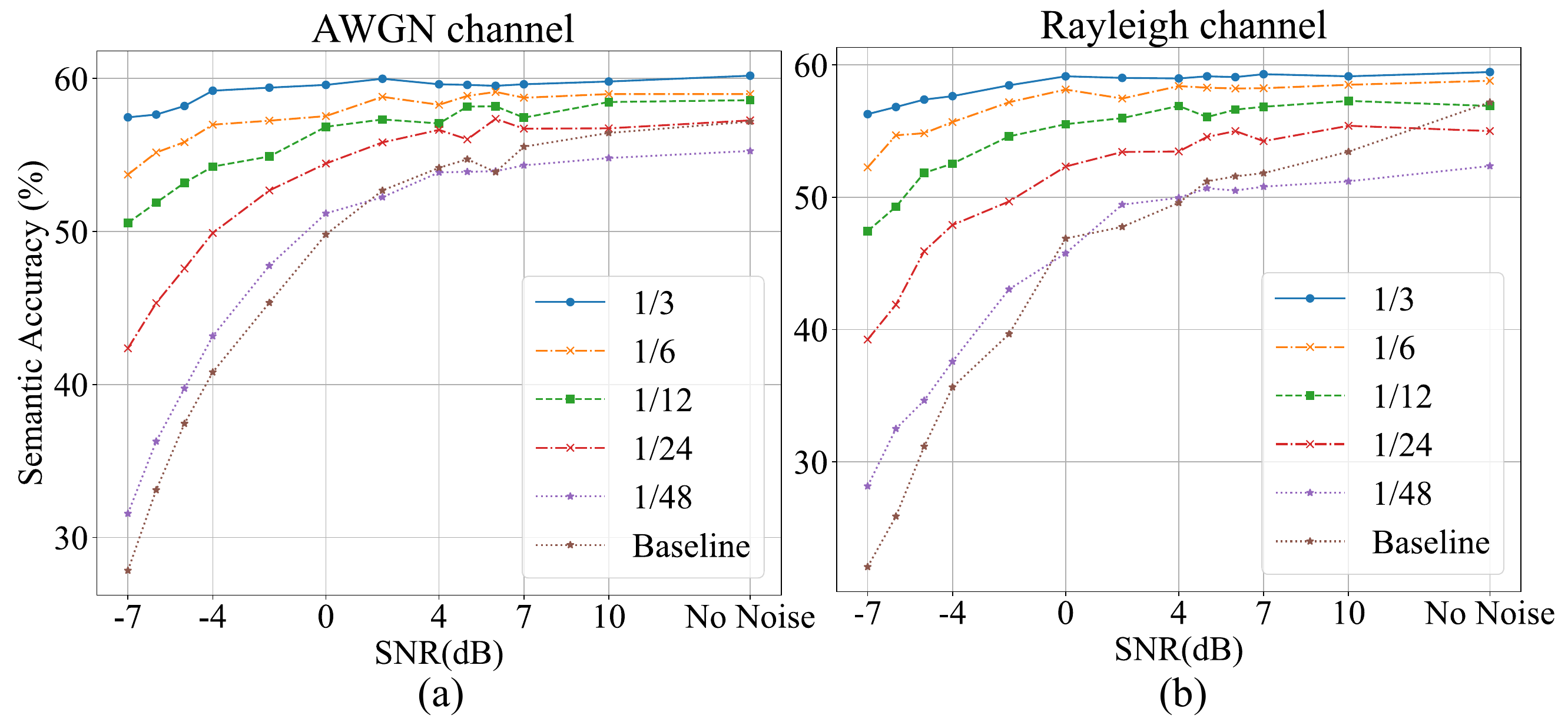}
	\caption{Comparison of semantic accuracy between our method and baselines under different CBRs}
	\label{fig_3}
\end{figure}

\subsection{Result Analysis}
The performance comparison of our proposed method with the baseline, which directly transmits CLIP features without compression, is illustrated in Fig.\ref{fig_3}. Our method consistently outperforms the baseline across most CBR settings. At CBR $=$ 1/48, the semantic accuracy is close to the baseline. However, our approach achieves this while transmitting only a 128-dimensional compressed feature, demonstrating the efficiency of the proposed feature compression method. As the CBR increases, our method continues to exhibit superior performance, confirming the benefits of learned compression and noise adaptation. As the SNR decreases, our method demonstrates a greater advantage over the baseline, highlighting its superior robustness in noisy environments. Furthermore, when CBR $>$ 1/12, the transmitted feature dimension exceeds 512, indicating that the network introduces additional redundancy into the features. This redundancy enhances noise resistance, enabling our method to maintain strong performance even in low-SNR conditions. Under Rayleigh channel, there is a slight decrease in the accuracy of all methods when compared to AWGN, but the overall trend remains the same.

\begin{figure}[h]
	\centering
	\includegraphics[width=3.5in]{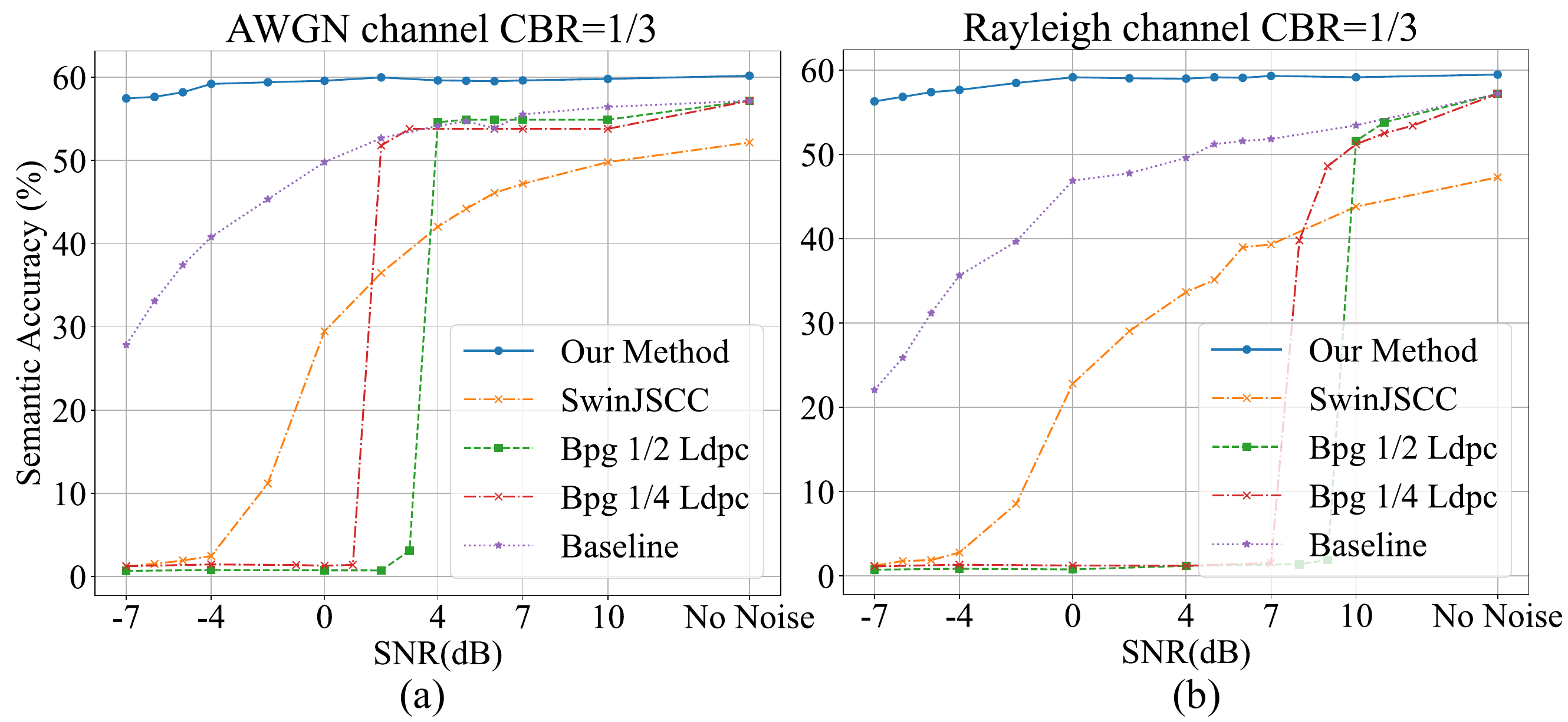}
	\caption{The comparison of semantic accuracy between our method and other methods in 1/3 CBR}
	\label{fig_4}
\end{figure}

The performance comparison of our proposed method with other methods in 1/3 CBR is shown in Fig.\ref{fig_4}. The Baseline method transmits a 512-dimensional feature vector, which corresponds to a CBR of 1/12. The performance of the traditional BPG+LDPC method depends on whether the channel decoder can correctly recover the source code at a given SNR. If decoding errors occur, the received image is distorted, as illustrated in Fig. 1(a), leading to a semantic accuracy of approximately 1\%. Once the SNR reaches a threshold where the channel decoder can fully recover the source code without errors, the semantic accuracy increases sharply to a normal level. The 1/4 LDPC scheme has stronger error correction capability due to its lower code rate, making it more robust to noise and capable of recovering the original image at lower SNR levels. SwinJSCC maintains a certain level of semantic accuracy at low SNRs but performs worse than BPG+LDPC at high SNRs. However, all these methods perform worse than the Baseline because the Baseline directly transmits semantic features, which inherently possess some degree of noise resistance. Our proposed method consistently achieves the best performance across all SNR levels, demonstrating its effectiveness in preserving semantic information while being more robust to noise.

\begin{table}[h]
	\caption{Speed comparison\label{tab:table1}}
	\centering
	\begin{tabular}{|c|c|}
		\hline
		method & inference time \\
		\hline
		Our & 5.7(CLIP)+1.0(Net)+1.2(KB)ms\\
		\hline
		SwinJSCC &  12.9ms\\
		\hline
	\end{tabular}
\end{table}

Table \ref{tab:table1} presents a comparison of inference speed. Our proposed method achieves a total inference time of 7.9 ms, which consists of 5.7 ms for CLIP feature extraction, 1.0 ms for the encoding and decoding network, and 1.2 ms for KB retrieval. In contrast, SwinJSCC requires 12.9 ms for inference. The significant reduction in inference time demonstrates the efficiency of our approach, which benefits from the lightweight encoder-decoder structure and the optimized FAISS-based retrieval process. This makes our method more suitable for real-time semantic communication applications, where low latency is crucial for practical deployment.

\section{conclusion}
This paper presented a KB-assisted and CLIP-based semantic image transmission framework that efficiently compresses and transmits semantic features while leveraging KB retrieval to select semantically similar images at the receiver. Experimental results demonstrate that our method can adapt to different transmission requirements by selecting an appropriate CBR. A lower CBR can be used to compress features and improve transmission efficiency, while a higher CBR allows for added redundancy, enhancing transmission robustness and performance. Our approach consistently outperforms than BPG+LDPC and SwinJSCC, demonstrating its effectiveness in preserving semantic information while maintaining robustness to channel noise. Future work will explore optimizing the KB structure and further improving feature compression techniques to enhance transmission efficiency in practical communication scenarios.

\bibliographystyle{IEEEtran}
\bibliography{IEEEabrv,KBSC}

\newpage

\vfill

\end{document}